\documentclass[journal]{IEEEtran}

\usepackage{cite}
\usepackage{amsmath,amssymb,amsfonts}
\usepackage{algorithmic}
\usepackage{graphicx}
\usepackage{flushend}
\usepackage{textcomp}
\usepackage{hyperref}
\usepackage{multirow,multicol}
\usepackage{pifont}% http://ctan.org/pkg/pifont
\newcommand{\cmark}{\ding{51}}%
\newcommand{\xmark}{\ding{55}}%
\usepackage[version=4]{mhchem}
\usepackage[table]{xcolor}

\definecolor{BrickRed}{RGB}{203, 65, 84}
\definecolor{ForestGreen}{RGB}{34, 139, 34}

\makeatletter
\newdimen\xfigwd
\makeatother

\usepackage[linesnumbered,ruled,vlined]{algorithm2e}

\def\BibTeX{{\rm B\kern-.05em{\sc i\kern-.025em b}\kern-.08em
    T\kern-.1667em\lower.7ex\hbox{E}\kern-.125emX}}

\begin{document}
% \history{Date of publication xxxx 00, 0000, date of current version xxxx 00, 0000.}
% \doi{10.1109/ACCESS.2023.DOI}

\title{ILASH: A Predictive Neural Architecture Search Framework for Multi-Task Applications}

\author{
   \IEEEauthorblockN{ \IEEEauthorrefmark{1}Md Hafizur Rahman, \IEEEauthorrefmark{2}Md Mashfiq Rizvee, \IEEEauthorrefmark{2}Sumaiya Shomaji} and \IEEEauthorrefmark{1}Prabuddha Chakraborty 
   
   \IEEEauthorblockA{
   \IEEEauthorrefmark{1}Department of Electrical and Computer Engineering, University of Maine\\
   \IEEEauthorrefmark{2} Department of Electrical Engineering and Computer Science, University of Kansas  } 
   % \\
   % Email: \IEEEauthorrefmark{1}\{md.hafizur.rahman, prabuddha\}@maine.edu} 
   % \IEEEauthorrefmark{2}\{mashfiq.rizvee,shomaji\}@ku.edu}

   }
   
\maketitle

% \tfootnote{This paragraph of the first footnote will contain support 
% information, including sponsor and financial support acknowledgment. For 
% example, ``This work was supported in part by the U.S. Department of 
% Commerce under Grant BS123456.''}

% \markboth
% {Author \headeretal: Preparation of papers for IEEE Access Journal}
% {Author \headeretal: Preparation of papers for IEEE Access Journal}
% \corresp{Corresponding author: Prabuddha Chakraborty (e-mail: prabuddha@maine.edu).}

% \titlepgskip=-15pt

\begin{abstract}
Artificial intelligence (AI) is widely used in various fields including healthcare, autonomous vehicles, robotics, traffic monitoring, and agriculture. Many modern AI applications in these fields are multi-tasking in nature (i.e. perform multiple analysis on same data) and are deployed on resource-constrained edge devices requiring the AI models to be efficient across different metrics such as power, frame rate, and size. For these specific use-cases, in this work, we propose a new paradigm of neural network architecture (ILASH) that leverages a layer sharing concept for minimizing power utilization, increasing frame rate, and reducing model size. Additionally, we propose a novel neural network architecture search framework (ILASH-NAS) for efficient construction of these neural network models for a given set of tasks and device constraints. The proposed NAS framework utilizes a data-driven intelligent approach to make the search efficient in terms of energy, time, and \ce{CO2} emission. We perform extensive evaluations of the proposed layer shared architecture paradigm (ILASH) and the ILASH-NAS framework using four open-source datasets (UTKFace, MTFL, CelebA, and Taskonomy). We compare ILASH-NAS with AutoKeras and observe significant improvement in terms of both the generated model performance and neural search efficiency with up to 16x less energy utilization, \ce{CO2} emission, and training/search time.
\end{abstract}


\section{Introduction}

\IEEEPARstart{A}{rtificial} intelligence (AI) is a rapidly growing market estimated to surpass 1 Trillion USD by 2027 \cite{bain2024aimarket}. AI is being used in most modern applications, devices, and services across almost all domains including automotive~\cite{weber2023approach,dong2023did,xu2023generative}, digital manufacturing~\cite{ponce2024developing,kusiak2024federated}, healthcare~\cite{chua2023tackling,bhardwaj2023enhanced, nandy2023intelligent}, and retail~\cite{wang2024embracing,vhatkar2024leveraging}. 
% The rapid increase in computation power of modern hardware has made running very large models possible paving a path towards Artificial general intelligence (AGI). 
However, one of the major concerns with AI, that can potentially slow down its growth, is rooted in the energy utilization and carbon emission aspects. 
In \cite{strubell2019energy}, researchers conducted a life cycle assessment on the training of various prevalent large AI models. The results indicated that this process could release over 626,000 pounds of carbon dioxide equivalent to almost five times the cumulative emissions of an average American car throughout its entire lifespan (including manufacturing). The two main sources of energy utilization and carbon emission are:
\begin{enumerate}
    \item \textbf{Neural Architecture Search (NAS) \& Training}: The automated NAS process can be extremely power-hungry. In certain cases, energy utilization can reach up to $4.98\times10^2$ kWh-PUE and carbon emission can exceed $4.75\times10^2$ lbs for just obtaining one AI model (see Table~\ref{tab:NAS_power}).
    \item \textbf{Inferencing}: AI models being used in billions of IoT edge devices (for inferencing tasks) also lead to a large amount of cumulative energy consumption and carbon emission. The ``\textit{State of IoT—Spring 2023}'' report shows a total of 14.3 billion active IoT endpoints~\cite{IoTAnalytics2023}.
\end{enumerate}

\begin{figure*}[ht]
\centering
\includegraphics[width=\linewidth]{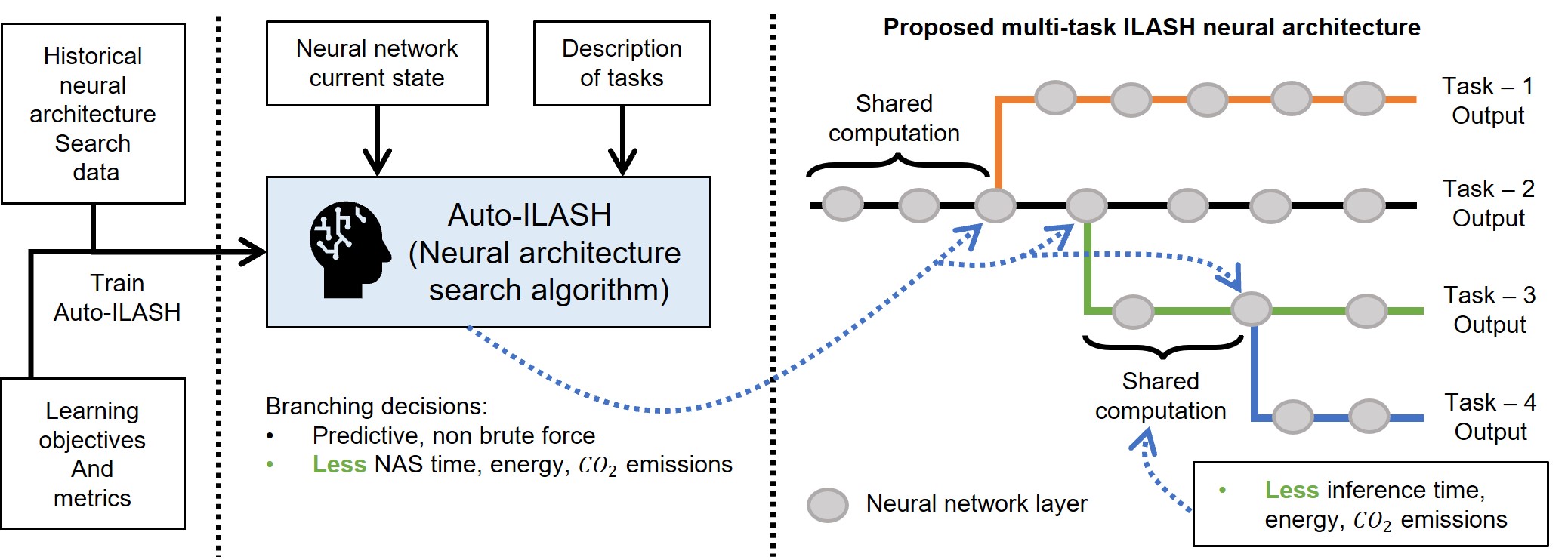}
\caption{Overview of the proposed ILASH architecture and the neural architecture search algorithm (ILASH-NAS).}
\label{fig:ILASH}
\end{figure*}

In this work, we propose to minimize both of these avenues of energy expenditure and \ce{CO2} emission by developing an \textbf{I}ntelligent \textbf{LA}yer \textbf{SH}ared (ILASH) neural architecture paradigm and an associated neural architecture search algorithm that leverages AI for efficiency. ILASH models reduce energy utilization and \ce{CO2} emission by sharing intermediate computation outcomes (layer-outputs) across different tasks (layer-sharing) while the proposed AI-Guided NAS (ILASH-NAS) makes building these ILASH models highly energy efficient. Our proposed ILASH architecture, the methodology of layer sharing, and the AI-guided neural architecture search process (for building ILASH networks) are novel compared to existing works.  
% The proposed AI-guided predictive NAS methodology build neural networks in with high efficiency and the 
% Additionally, we make inferencing more energy efficient (and less \ce{CO2} emitting), and we use ILASH to create neural architectures that can perform multiple tasks leveraging a large amount of common computations. 
We use ILASH-NAS to search models for different multi-task applications using standard datasets such as UTKFace \cite{das2018mitigating}, MTFL \cite{zhang2014facial}, CelebA \cite{liu2018large}, and Taskonomy \cite{zamir2018taskonomy} with Raspberry Pi 4 Model B, Jetson Orin Nano, Jetson Nano, and Jetson AGX nano as target edge devices. We observe up to 16x reduction in energy utilization and \ce{CO2} emission during the neural architecture search and training process compared to state-of-the-art multi-task NAS algorithms (AutoKeras). The layer-sharing aspect of the final model also led to 3x reduction in inferencing energy and \ce{CO2} emission compared to state-of-the-art counterparts. In this article, we make the following research contributions:
\begin{enumerate}
    \item Designed a novel class of neural architectures (ILASH) for multi-task applications that reduces overall computation by sharing layers among tasks leading to reduced energy/emission overheads during inferencing.
    \item Formulated a novel neural architecture search algorithm, appropriate for building ILASH models (ILASH-NAS), that uses AI to boost search efficiency while reducing energy/emission overheads. 
    \item Implemented a highly parameterized version of the ILASH-NAS that can generate layer-shared architectures for multi-task applications.
    \item Evaluated ILASH-NAS and the generated ILASH models using standard datasets such as UTKFace, MTFL, CelebA, and Taskonomy on different edge devices (NVIDIA Jetson AGX Orin, Orin Nano, Jetson Nano, and Raspberry Pi 4 Model B). 
\end{enumerate}

% The subsequent sections of the paper are structured as follows: Section 2 provides background and motivation; Section 3 outlines the methodology employed in this study; Section 4 addresses the experiments conducted and the data description; Section 5 analysis the experimental results; and, lastly, Section 6 presents our concluding remarks.

%According to the Embedded survey 2023\cite{nitin2023} embedded projects targeting a wide range of applications: 24\% of IoT, 29\% of industrial automation, 17\% of AI or embedded AI. The latest IoT Analytics “State of IoT—Spring 2023” report shows that the number of global IoT connections grew by 18\% in 2022 to 14.3 billion active IoT endpoints. In 2023, IoT Analytics expects the global number of connected IoT devices to grow another 16\%, to 16.7 billion active endpoints \cite{Satyajit2023}.

% There was an estimated 15.14 billion connected IoT devices in 2023, with a forecast of 29.42 billion devices by 2030 \cite{Lionel2023}.

\section{Background and Motivation}
In this section, we will briefly describe multi-task learning and neural architecture search (NAS) algorithms.

% This approach entails instructing the model to generate indistinguishable characteristics from single-task networks through the process of minimizing a linear combination of losses relevant to each task. The fundamental focus of MTL lies in reducing the size of the network while minimize the loss jointly. In one research, 
% Yuan \emph{et al.} \cite{gao2019nddr} proposed the NDDR-CNN Network which is an encoder-centric MTL model utilizing automatic feature fusion across various tasks at each layer. 

\begin{table}[ht]
\renewcommand{\arraystretch}{1.3}
\scriptsize\addtolength{\tabcolsep}{-3pt}
\centering
\caption{Comparison of energy consumption and \ce{CO2} emission among state-of-the-art NAS methods on CIFAR-10 (\textit{Not a Multitask application}).}
\label{tab:NAS_power}
\begin{tabular}{ccccc}
\hline
\textbf{Method} &
\textbf{\begin{tabular}[c]{@{}c@{}}NVIDIA \\ 
    GPU \end{tabular}} &
  \textbf{GPU Hours} &
  \textbf{kWh-PUE} &
  \textbf{\begin{tabular}[c]{@{}c@{}}\ce{CO2} emission \\ (lbs)\end{tabular}} \\ \hline

AutoSNN {\cite{na2022autosnn}}   & RTX 2080 Ti & $4.35\times10^1$ & $2.58\times10^1$ & $2.46\times10^1$ \\ 

DCNN {\cite{ma2020autonomous}} & Tesla TitanXp  & $2.40\times10^2$ & $1.42\times10^2$ & $1.36\times10^2$ \\ 

CNN-GA {\cite{sun2020automatically}} & GTX 1080 Ti  & $8.40\times10^2$ & $4.98\times10^2$ & $4.75\times10^2$ \\ 

NAS-RL {\cite{zoph2016neural}} & Tesla K40                                                       & $5.38\times10^5$ & $3.12\times10^5$ & $2.98\times10^5$ \\ 

EAS {\cite{cai2018efficient}} & GTX 1080                                                    & $2.40\times10^2$ & $1.02\times10^2$ & $9.77\times10^1$ \\

CGP-CNN{\cite{suganuma2020evolution}}  & GTX 1080
& $6.48\times10^2$ & $2.76\times10^2$ & $2.63\times10^2$ \\ 

DARTS-PT{\cite{wang2021rethinking}}  & GTX 1080 Ti
& $19.2$ & $8.19$ & $7.81$ \\
NAO {\cite{luo2018neural}} & Tesla V100                                                       & $4.80\times10^3$ & $3.41\times10^3$ & $3.26\times10^3$ \\ 
PNAS {\cite{liu2018progressive}} & Tesla P100                                                      & $5.40\times10^3$ & $3.20\times10^3$ & $3.05\times10^3$ \\ 

NASNet {\cite{zoph2018learning}}   & Tesla P100                                                     & $4.80\times10^4$ & $2.84\times10^4$ & $2.71\times10^4$ \\ 

Hierarchical-EAS {\cite{liu2017hierarchical}} & Tesla P100  & $7.20\times10^3$ & $4.27\times10^3$ & $4.07\times10^3$ \\ \hline

\end{tabular}
% \vspace{-0.15in}
\end{table}

\begin{table*}[ht]
\renewcommand{\arraystretch}{1.3}
\scriptsize\addtolength{\tabcolsep}{1pt}
\centering
\caption{Qualitative comparison between ILASH-NAS and other neural architecture search frameworks.}
\label{tab:NAS_compare}
\begin{tabular}{cccccccc}
\hline
\multirow{2}{*}{\textbf{Method}} &
 \multirow{2}{*}{ \textbf{\begin{tabular}[c]{@{}c@{}}Efficiency based\\  on GPU usage\end{tabular}}} & 
  \multirow{2}{*}{ \textbf{\begin{tabular}[c]{@{}c@{}}Search Strategy\end{tabular}}} & 
 \multicolumn{3}{c}{ \textbf{Multi-task  Support}} & 
 \multirow{2}{*}{ \textbf{\begin{tabular}[c]{@{}c@{}}Single task \\ Support\end{tabular}}} &
  \multirow{2}{*}{ \textbf{AI Driven}} \\ \cline{4-6}
 & & & \textbf{Regression} & \textbf{Classification} & \textbf{2D Semantic} & \\ \hline
  
NAS-RL{\cite{zoph2016neural}}                                                   & Very Low  & Cell-based  & \color{BrickRed}{\Large \xmark} & \color{BrickRed}{\Large \xmark}  &  \color{BrickRed}{\Large \xmark} &     \color{ForestGreen}{\Large \cmark} & \color{BrickRed}{\Large \xmark} \\
EAS {\cite{cai2018efficient}}                                                           & High & Transformation based & \color{BrickRed}{\Large \xmark} & \color{BrickRed}{\Large \xmark}  &  \color{BrickRed}{\Large \xmark} &   \color{ForestGreen}{\Large \cmark} & \color{BrickRed}{\Large \xmark} \\ 
Hierarchical-EAS {\cite{liu2017hierarchical}} & Medium & Hierarchical &  \color{BrickRed}{\Large \xmark} & \color{BrickRed}{\Large \xmark}  &  \color{BrickRed}{\Large \xmark} &    \color{ForestGreen}{\Large \cmark} &\color{BrickRed}{\Large \xmark} \\ 

AmoebaNet-A  {\cite{real2019regularized}}\ & Low   &  Cell-based  & \color{BrickRed}{\Large \xmark} & \color{BrickRed}{\Large \xmark}  &  \color{BrickRed}{\Large \xmark}  &    \color{ForestGreen}{\Large \cmark} & \color{BrickRed}{\Large \xmark} \\ 

DARTS {\cite{liu2018darts}} & Medium &   Cell-based  & \color{BrickRed}{\Large \xmark} & \color{BrickRed}{\Large \xmark}  &  \color{BrickRed}{\Large \xmark}  &     \color{ForestGreen}{\Large \cmark} & \color{BrickRed}{\Large \xmark} \\ 

PNAS{\cite{liu2018progressive}}                                                     & Low  & Cell-based   & \color{BrickRed}{\Large \xmark} & \color{BrickRed}{\Large \xmark}  &  \color{BrickRed}{\Large \xmark}  &     \color{ForestGreen}{\Large \cmark} & \color{BrickRed}{\Large \xmark} \\ 

%%%%%%%%%%%%%
AutoSNN {\cite{na2022autosnn}}                                                     & Medium  & Spiking Neural Network based  &  \color{BrickRed}{\Large \xmark} & \color{BrickRed}{\Large \xmark}  &  \color{BrickRed}{\Large \xmark}  &    \color{ForestGreen}{\Large \cmark} & \color{BrickRed}{\Large \xmark} \\ 

Autokeras {\cite{jin2019auto, JMLR:v24:20-1355}}                                                        & Medium & Network Morphism &  \color{ForestGreen}{\Large \cmark} & \color{ForestGreen}{\Large \cmark}  &  \color{BrickRed}{\Large \xmark}   &   \color{ForestGreen}{\Large \cmark} & \color{BrickRed}{\Large \xmark} \\ 

CNN-GA {\cite{sun2020automatically}} & Low  & Mutation and Crossover  & \color{BrickRed}{\Large \xmark} & \color{BrickRed}{\Large \xmark}  &  \color{BrickRed}{\Large \xmark}   &     \color{ForestGreen}{\Large \cmark} & \color{BrickRed}{\Large \xmark} \\ 

DARTS-PT{\cite{wang2021rethinking}} & High & Gradient Stabilazation  & \color{BrickRed}{\Large \xmark} & \color{BrickRed}{\Large \xmark}  &  \color{BrickRed}{\Large \xmark}  &    \color{ForestGreen}{\Large \cmark} & \color{BrickRed}{\Large \xmark} \\ 

GENIUS {\cite{zheng2023can}}                                                   & Medium & NAS-Bench  & \color{BrickRed}{\Large \xmark} & \color{BrickRed}{\Large \xmark}  &  \color{BrickRed}{\Large \xmark}  &  \color{ForestGreen}{\Large \cmark} & \color{ForestGreen}{\Large \cmark} \\ 

DCNN {\cite{ma2020autonomous}}                                                   & Low  & Mutation and Crossover  & \color{BrickRed}{\Large \xmark} & \color{BrickRed}{\Large \xmark}  &  \color{BrickRed}{\Large \xmark}   &     \color{ForestGreen}{\Large \cmark} & \color{BrickRed}{\Large \xmark} \\ 

CGP-CNN{\cite{suganuma2020evolution}}  & Low & Mutation based &  \color{BrickRed}{\Large \xmark} & \color{BrickRed}{\Large \xmark}  &  \color{BrickRed}{\Large \xmark}   &    \color{ForestGreen}{\Large \cmark} & \color{BrickRed}{\Large \xmark} \\ 

LeMo-NADe{\cite{rahman2024lemo}}  &  Medium  & Open Search  & \color{BrickRed}{\Large \xmark}  & \color{BrickRed}{\Large \xmark} &  \color{BrickRed}{\Large \xmark}   & \color{ForestGreen}{\Large \cmark} & \color{ForestGreen}{\Large \cmark} \\ \hline

\textbf{ILASH-Heu (Proposed)}                                                           & \textbf{Medium} & Branching based &  \color{ForestGreen}{\Large \cmark} & \color{ForestGreen}{\Large \cmark} & \color{ForestGreen}{\Large \cmark}  &    \color{ForestGreen}{\Large \cmark} & \color{BrickRed}{\Large \xmark} \\ 

\textbf{ILASH-Pred (Proposed)}                                                           & \textbf{Very High } & Branching based &  \color{ForestGreen}{\Large \cmark} & \color{ForestGreen}{\Large \cmark} & \color{ForestGreen}{\Large \cmark}  &    \color{ForestGreen}{\Large \cmark} & \color{ForestGreen}{\Large \cmark} \\ \hline
\end{tabular}
% \vspace{-0.15in}
\end{table*}

\subsection{Traditional Neural Architecture Search}
Neural Architecture Search (NAS) algorithms are widely used for finding optimal neural architectures for a given application. Recently, Spiking Neural Networks (SNNs) have garnered significant attention, with studies like AutoSNN by Na \emph{et al.} \cite{na2022autosnn}, which explore hyperparameters within a pre-established search space. SeqNAS was introduced by Udovichenko \emph{et al.} \cite{udovichenko2024seqnas} to classify event sequences, using Bayesian optimization in conjunction with multi-head self-attention convolutional neural networks. Meanwhile, Wang \emph{et al.} proposed DARTS-PT \cite{wang2021rethinking}, a perturbation-based approach that improves generalization through the evaluation of supernet operation. Even with all these innovations, state-of-the-art (SOTA) NAS algorithms are still iterative in nature, resource-intensive \cite{rizvee2023understanding} and time-consuming (see Table~\ref{tab:NAS_compare} and Table~\ref{tab:NAS_power}). Moreover standard NAS techniques are not designed for building multi-task neural networks and hence a direct comparison with ILASH is not possible.
% However, a drawback is that the optimization process may not consistently converge to the optimal solution, highlighting the critical importance of selecting an appropriate learning rate. Moreover, DARTS employs a cell-based search space, where the efficiency variance is minimal. This could be a limiting factor, as sophisticated search techniques may only yield marginal improvements over the mean performance of randomly selected structures. Additionally, the application of gradient-based methods requires expertise in supernet building before utilization.
% Although NAS build AI model automatically it takes longer computing time and energy to find the optimal architecture for the given problem. 
Table~\ref{tab:NAS_power} shows the statistics of GPU hours, energy consumption, and equivalent \ce{CO2} emission associated with different NAS methods for the CIFAR10 dataset. Energy consumption and \ce{CO2} emission are calculated using~\cite{strubell2020energy}:

\vspace{-0.1in}
\begin{equation}
p_t=\frac{1.58 t\left(p_c+p_r+g p_g\right)}{1000}
\label{eqn1}
\end{equation}

\vspace{-0.1in}

\begin{equation}
\ce{CO2} \mathrm{e}=0.954 p_t
\label{eqn2}
\end{equation}
% \vspace{-0.1in}

Here, $t$, $p_c$, $p_r$, and $p_g$ denote the NAS runtime, average power consumed by the CPU, verage power consumed by RAM, and verage power consumed by GPU, respectively. `g' stands for the total number of GPUs, and $p_t$ is the power usage effectiveness (PUE) measured in kilowatt-hours (kWh). We assume the power usage of GTX 1080, GTX 1080 Ti, RTX 2080 Ti, and Tesla TitanXp to be 270 watt, 375 watt, 375 watt, and 375 watt respectively ($1.5\times TDP$ \cite{cutress2022intel}). 

\subsection{Multi-Task Learning and NAS}
Multi-task learning involves training a single Artificial Intelligence (AI) model for performing multiple tasks. Multi-task learning is typically done to reduce weight size overhead for IoT devices.
% with limited storage space for diverse applications such as object detection and semantic segmentation. 
Recently, works such as Fast GraspNeXt \cite{wong2023fast}, used a self-attention neural network model designed specifically for embedded multi-task learning. Li \emph{et al.} \cite{li2020knowledge} proposed a multi-task learning (MTL) strategy utilizing knowledge distillation to attain fair parameter sharing among multiple tasks. However, these techniques do not focus on building multi-task neural networks for reducing overall energy consumption and \ce{CO2} emissions. Hence a direct comparison of ILASH with these techniques is not possible. However we qualitatively compare different NAS techniques with ILASH-NAS in Table~\ref{tab:NAS_compare}. 
Auto-Keras, a framework developed by Jin et al. in 2019 \cite{jin2019auto}, implements some of the most well-known NAS algorithms such as Greedy, Random, and Bayesian in the context of multi-task learning (only publicly available framework). 
% It is also the only publicly available framework that supports multi-task learning (with  hard-parameter sharing). 
Hence, we compare ILASH with Auto-Keras algorithms to understand the effectiveness of ILASH (Sec~\ref{sec:res}). 

\begin{algorithm}[!ht]
\DontPrintSemicolon % Some LaTeX compilers require you to use \dontprintsemicolon instead
\KwIn{$Model_{base},Dataset,TaskInfoDict, ll, ul$}
\KwOut{$ILASH\_Model, ILASH\_Dataset$}

$ILASH\_Dataset=[ ]$; 
$ILASH\_Model \gets \emptyset$ \\
\For{$taskIdx \gets 0$ \textbf{to} $len(TaskInfoDict)$} 
{   
     \If {$taskIdx=0$}
     {
        $ILASH\_Model \gets Model_{base}$ \\
        $Data \gets Dataset[taskIdx]$ \\
        $ILASH\_Model.train(Data.tr)$ 
        
     }
     \Else
     {
        $Model \gets ILASH\_Model $ \\
        $Data \gets Dataset[taskIdx]$;
        $BGN \gets 0$ \\
        \For {$idx, lr \in Model.layers()$}
        {
            \If {$(idx < ll)$ OR $(idx > ul)$}
            {
                $continue$
            }
            $temp\_model \gets branch(Model,lr) $\\
            $temp\_model.train(Data.tr)$ \\
            $GN \gets Eval(temp\_model, Data.val)$

            \If {$GN>BGN$}
            {
            $BGN \gets GN$ \\
            $ILASH\_Model \gets temp\_model$
            }

            $taskInfo \gets TaskInfoDict[taskIdx]$\\
            $layerInfo \gets encode(idx, Model)$\\
            $D \gets [taskInfo,layerInfo,GN]$ \\
            $ILASH\_Dataset.append(D) $

        }

     }
}

\Return{$ILASH\_Model, ILASH\_Dataset$}
\caption{Heuristic Based Search: ILASH-Heu}
\label{algo:Heuristic_Based_Search}
\end{algorithm}
\begin{algorithm}[ht]
\DontPrintSemicolon % Some LaTeX compilers require you to use \dontprintsemicolon instead
\KwIn{$Model, idx$}
\KwOut{$encodedLayer$}

$temp \gets \emptyset$ \\

\For{$i \gets -1$ \KwTo $1$}{

    $lrType \gets Model.layers[idx+i].type$\\
    
    \If {$lrType==Conv2D$ OR $DepthwiseConv2D$}
    {
    $lr \gets Model.layers[idx + i]$ \\
    $temp.kernel_i \gets lr.kernelSize$ \\
   $temp.pad_i \gets (1 \text{ if } lr.pad == valid \text{ else } 0)$ \\
    $temp.stride_i \gets lr.strideSize$ \\
    }

    \Else
    {
        $temp.kernel_i \gets -1$ \\
        $temp.pad_i  \gets -1$ \\
        $temp.stride_i  \gets -1$ \\
    }
}

$encodedLayer \gets temp$ \\
\Return{$encodedLayer$}\;
\caption{Layer Encoding: encode}
\label{algo:encoder}
% \vspace{-0.15in}
\end{algorithm}

\section{Methodology}
Next we formalize the ILASH architecture and the ILASH neural architecture search algorithms. 

\subsection{ILASH: Layer Shared Architecture}
Fig.~\ref{fig:ILASH} shows the overview of our proposed ILASH architecture. The proposed approach is a hybrid between the hard-layer-sharing approaches and the non-sharing approach (i.e. separate models for each task). Assume that a system needs to perform $p$ tasks ($T_1$, $T_2$, ..., and $T_P$) such that $T_1 = \{l_1^1, l_2^1, l_3^1, ...,l_{n1}^1\}$, $T_2 = \{l_1^2, l_2^2, l_3^2, ...,l_{n2}^2\}$, and $T_P = \{l_1^p, l_2^p, l_3^p, ...,l_{np}^p\}$. Here $l_x^y$ is the $x^{th}$ layer of the $y^{th}$ task. Also $P$ is a set of numbers \{$1, 2, ...,  p$\}. The differences between these strategies are as follows.
\begin{enumerate}
    \item \textbf{No Sharing:} For this, $\forall x \in T_a$, $x\not\in$ $\bigcup_{i \in P-a}T_i$. This implies that, no two layers are the same among all the tasks (i.e. completely separate models).
    \item \textbf{Hard Layer Sharing:} In this case, $\forall y \in P $, $l_1^y$ are the same, $l_2^y$ are the same, and so on up to the second last layer. Only the last layer is different across all tasks.
    \item \textbf{ILASH [\textit{Proposed}]:} First we build $T_1 = \{l_1^1, l_2^1, ...,l_{n1}^1\}$. Next we identify a layer $x \in T_1$ for branching out to support Task 2. This will lead to  $T_2 = \{l_1^1, l_2^1, l_3^1, ..., l_x^1, l_{x+1}^2, ...,l_{n2}^2\}$. Note that we are reusing layers $\{l_1^1, l_2^1, l_3^1, ..., l_x^1\}$ from Task 1 to support Task 2. Next to support Task 3, we pick a branching layer from the set of all unique layers ($T_1\cup_{}T_2$), connect extra layers from that branching point to support task 3, and reuse all prior layers from that point for task 3. This process continues until all tasks are supported. Fig.~\ref{fig:ILASH} provides a visual representation of the proposed network architectures. The ILASH architecture is a hybrid between `\textit{No Sharing}' and \textit{`Hard Layer Sharing'}.
\end{enumerate}

For ILASH, the branching process determines the efficacy for the overall network and to efficiently handle this process, we propose two mechanisms for ILASH neural architecture search (ILASH-NAS) described next: (1) A heuristics approach and (2) A predictive approach using AI/ML. 

\subsection{Heuristic Based ILASH-NAS}
Our initial goal is to build a heuristic-driven neural network search algorithm for layer-shared models that will allow us to capture the necessary information (dataset) required to build an AI-guided version of the NAS framework. This NAS algorithm (ILASH-Heu) is described using Algorithm~\ref{algo:Heuristic_Based_Search} and Algorithm~\ref{algo:encoder}. 

%This algorithm takes as input: (1) A base model ($Model_{base}$) which will be referenced for building the final ILASH model; (2) the $Dataset$ containing training/validation data of a given multi-task application; (3) A dictionary containing information about all the tasks ($TaskInfoDict$); (4) a lower limit ($ll$) and (5) a upper limit ($ul$). The lower and upper bounds determine the layer position boundaries at which branching needs to occur. Essentially, these boundaries define the range within the structure or model, defining the beginning and end locations where the branching process should take place.

The inputs of Algorithm~\ref{algo:Heuristic_Based_Search} are as follows:

\begin{itemize}
    \item $Model_{base}$: This is the neural network model upon which all subsequent adaptations and refinements will be made. It acts as the model for Task-1.
     \item $Dataset$: The dataset is the collection of data points used for training the model (for all tasks). It includes input data and corresponding target labels, providing the necessary information for the model to learn and adapt.
     \item $TaskInfoDict$: This dictionary contains information about each task such as input shape of the data, shape of the expected output, choice of loss function, metrics used for evaluating the model's performance, and the hyperparameters to fine-tune the model during training.
    \item Upper limit ($ul$) and Lower limit ($ll$): The lower and upper limits determine the layer position boundaries between which branching is allowed.

 \end{itemize}

% In this method we require a base model to which ILASH branch its layer based on the given task. 
% Algorithm \ref{algo:Heuristic_Based_Search} serves as a detailed blueprint for a heuristic-based ILASH search process. At the heart of this algorithm is the Heuristic ILASH module, which acts as the central piece of the puzzle.

\begin{algorithm}[ht]
\DontPrintSemicolon

\KwIn{\parbox[t]{0.5\textwidth}{
    $Model_{base}, Auto\_ILASH, Dataset, TaskInfo, \\
    ll, ul$
}}
%\KwIn{$[Model_{base}, Auto-ILASH, Dataset, TaskInfo, ll, ul]$}
\KwOut{$BestModel$}
$ILASH\_Model \gets \emptyset$ \\
\For{$T\_idx \gets 0 $ \textbf{to} $len(TaskInfo)$} 
{   
    \If {$T\_idx = 0$}
    {
        $ILASH\_Model \gets Model_{base} $ 
        
    }
    \Else
    {
        % $Model \gets ILASH\_Model $ \\
        $BGN \gets 0 $;
        $bestBr \gets \emptyset $\\
        \For {$idx, lr \in ILASH\_Model.layers()$}
        {
            \If {$(idx < ll)$ OR $(idx > ul)$}
            {
                $continue$\\
            }
            $data \gets [TaskInfo[T\_idx], lr.encode()]$ \\
            $GN\gets Auto\_ILASH.predict(data)$ \\
            \If {$GN>BGN$}
            {
                $BGN \gets GN$ \\
                $bestBr \gets lr$\\
            }

        }
        $ILASH\_Model.branch(bestBr)$

    }

}
$ILASH\_Model.train(Dataset.tr)$ \\

\Return{$ILASH\_Model$}\;
\caption{Predictive Based Search: ILASH-Pred}
\label{algo:Predictive_Based_Search}
\end{algorithm}

In Algorithm \ref{algo:Heuristic_Based_Search}, ILASH-Heu initiates the process by training the base model on the task-1 dataset (lines 3 to 6). From the second task onward, the algorithm attempts to branch off from all layers having a depth higher than $ll$ and lower than $ul$ in the existing network (line 11). Based a branching choice/location ($lr$) a temporary network (temp\_model) is created by replicating the base model layers from the branching location ($lr$) to the $2^{nd}$ last layer. An output layer based on the number of outputs of the task is added at the end with an appropriate activation function depending on the type of task (classification, regression, etc.). In lines 14 and 15, this temporary network (temp\_model) is then trained and evaluated for obtaining its Goodness Metric ($GN$) using Eqn~\ref{eq:metric}. The model (ILASH\_Model) corresponding to the best branching location ($lr$) and the best Goodness Metric value ($BGN$) are tracked in lines 16-18. Inside ILASH\_Dataset (line 22), we track the Goodness Metric ($GN$) value associated with each branching decision along with information of the task being handle and an encoding computed based on the layer from which branching was done. Algorithm \ref{algo:Heuristic_Based_Search} returns a layer-shared model for all the given tasks in line 23. Algorithm \ref{algo:encoder} shows the overall layer encoding process which takes the $Model$ and branching layer index ($idx$) as input and generates an encoding for representing this branching layer. The encoding process captures the information about the branching layer and its surrounding layers as a vector representation. This encoding is required later for the predictive ILASH-NAS process (discussed in Sec.~\ref{sec:predIlash}). 

% returns the encoded layer where previous, current, and next layers information are encoded. In lines 4-8, layer information (Number of Kernel, kernel size, padding, and stride) is encoded for Conv2D and DepthwiseConv2D layers. Kernel size, padding, and stride number are set to -1 for other layers (batch norm, relu) for encoding purpose.

% Based on the outcome of this check, ILASH strategically performs layer branching to adapt the model architecture for enhanced performance on the subsequent task. Layer branching function is mainly done by replicating the base model layer from the branching location ($lr$) to the $2^{nd}$ last layer of the model. An output layer based on the number of outputs of the task is added at the end with an appropriate activation function. 
%To ensure the quality of the model we also proposed a metric called Goodness (GN) which can be calculated from the following equation~\ref{eq:metric}.

To address the concern regarding how energy consumption and carbon emissions are considered (or the lack thereof) in the NAS process, we propose a metric called Goodness (GN). This metric is calculated using the following equation:

\begin{equation}
GN = acc \times (1-G_{th}) + \left(\frac{lr_{index}}{lr_{total}}\right) \times G_{th}
\label{eq:metric}
\end{equation}

Here, $acc$ represents the validation accuracy of the model, while the green threshold ($G_{th}$) is a user-controlled parameter that balances accuracy and network efficiency. The term $\frac{lr_{index}}{lr_{total}}$ indicates the relative position of the layer within the total number of layers, with $G_{th}$ serving as a weight to reflect the preference between accuracy and efficiency.
By integrating $lr_{index}/lr_{total}$ into the GN metric, we implicitly consider network efficiency, which is closely related to energy consumption and carbon emissions.
% Models with fewer parameters and layers, typically associated with lower energy consumption, are favored when the $G_{th}$ value increases.

%Here, $acc$ represents the validation accuracy of the model, green threshold ($G_{th}$) is a user-controlled parameter that can be used to tradeoff between accuracy and network efficiency, the term $lr_{total}/lr_{index}$ represents the relative position of the layer within the total layers, and $G_{th}$is applied as a weight in this context. The overall result, GN, provides a combined metric that incorporates accuracy and a preference for fewer model parameters based on layer position.
% In lines 16-18, we compare the Goodness values and update the ILASH\_model.
% This process is continued for all the tasks and finally, the algorithm returns a layer-shared model for all the given tasks. 
% In lines 19-22, we collect task information, layer encoding, and corresponding goodness metrics into the ILASH benchmark dataset (ILASH\_Dataset).

% information to make the ILASH benchmark dataset (ILASH\_Dataset) which contains different layers' information with dataset features and corresponding goodness metrics. 

\subsection{Predictive ILASH-NAS}
\label{sec:predIlash}
The first step of the predictive ILASH\_NAS is to train an AI model ($Auto-ILASH$) that can predict the goodness (GN) metric for a given branching choice during the ILASH model search. This prediction significantly speeds up the search process as demonstrated in the results section. We have experimented with different machine learning (ML) algorithms (to train the $Auto-ILASH$ model) and we measure the performance of these ML algorithms using the following metrics.

\begin{itemize}
    \item \textbf{Mean Absolute Error (MAE)}: We can calculate MAE as follows:
    \begin{equation}
        \text{MAE} = \frac{1}{n} \sum_{i=1}^{n} \left| y_i - \hat{y}_i \right|
    \end{equation}
    Here, n is the number of data points, $y_i$ is the actual value, $\hat{y}_i$ is the predicted value and $\left| y_i - \hat{y}_i \right|$ is the absolute error for each data point.

    \item \textbf{Mean Squared Error (MSE)}: We calculate the MAE using the following equation:
    \begin{equation}
        \text{MSE} = \frac{1}{n} \sum_{i=1}^{n} (y_i - \hat{y}_i)^2
    \end{equation}

    Here, n is the number of data points, $y_i$ is the actual value, $\hat{y}_i$ is the predicted value and $(y_i - \hat{y}_i)^2$ is the squared error for each data point.

    \item \textbf{Root Mean Squared Error (RMSE)}: The mean root squared error (RMSE) is calculate as follows:
    \begin{equation}
        \text{RMSE} = \sqrt{\frac{1}{n} \sum_{i=1}^{n} (y_i - \hat{y}_i)^2}
    \end{equation}

    Here, n is the number of data points, $y_i$ is the actual value, $\hat{y}_i$ is the predicted value and $(y_i - \hat{y}_i)^2$ is the squared error for each data point.

    \item \textbf{$R^2$ Score}: We calculate the $R^2$ score using the following equation: 

    \begin{equation}
        R^2 = 1 - \frac{\sum_{i=1}^{n} (y_i - \hat{y}_i)^2}{\sum_{i=1}^{n} (y_i - \bar{y})^2}
    \end{equation}

    Here, n is the number of data points, $y_i$ is the actual value, $\hat{y}_i$ is the predicted value and $\bar y$ is the mean of actual values.

\end{itemize}

Algorithm \ref{algo:Predictive_Based_Search} shows the predictive ILASH-NAS process. Unlike ILASH-Heu, ILASH-Pred does not need to train the base model on the first given task. For each subsequent task, the Best Goodness (BGN) value is set to 0 and the best branching (bestBr) is set to empty in lines 6. In lines 7-13, we search for the best layer for branching and perform the branching process. Before predicting the Goodness (GN) of each layer we first check if the layer is in between $ul$ and $ll$. In line 11, $Auto-ILASH$ predicts the GN of the layer by using the encoding computed in line 10. In lines 12-14, we keep track of the best branching point ($bestBr$) and the best Goodness metric value ($BGN$). After checking all branching layer possibilities, ILASH-Pred builds a branched model for the task using $bestBr$ (line 15). Based the branching choice/location ($bestBr$) the branched model is created by replicating the base model layers from the branching location ($bestBr$) to the $2^{nd}$ last layer. An output layer based on the number of outputs of the task is added at the end with an appropriate activation function depending on the type of task (classification, regression, etc.). This process is continued for the remaining tasks. The final model ($ILASH\_Model$) is trained on the training data for all tasks at the same time.

\section{Experimental Setup and Data Description}
Next we will discuss the experimental setup and the datasets.
\label{sec:res}
\subsection{Experimental Setup: Auto-ILASH}
Auto-ILASH is the predictive component of the ILASH-Pred version of ILASH-NAS (see Algorithm~\ref{algo:Predictive_Based_Search}. Auto-ILASH is an ML model and we have experimented with the following ML algorithms for realizing it (evaluation results in later section).
\begin{itemize}
    \item \textbf{Random Forest (RF) Regressor}: Number of estimator = 100, criterion = `gini', max\_features = sqrt, max\_features = 1.0, min\_samples\_split = 2.
    \item \textbf{Decision Tree (DT) Regressor}: Criterion = `squared\_error', splitter = `best', min\_sample\_split = 2. 
    \item \textbf{Linear Regressor}: For Linear Regressor, we used copy\_X = True, fit\_intercept = True.
    \item \textbf{Support Vector Regressor (SVR)}: For SVR, we used kernel = `rbf', gamma = `scale', and epsilon = 0.1.
    \item \textbf{Gradient Boosting (GradBoost) Regressor}: For GradBoost Regressor, we used `squared\_error' loss, learning rate = 0.1, number of estimator = 100 and criterion = `friedman\_mse’.
\end{itemize}

We use a leave-one-out strategy for creating the Auto-ILASH model. For example, when searching for the ILASH network for UTKFace, the Auto-ILASH model used by Algorithm~\ref{algo:Predictive_Based_Search} is trained on the two merged ILASH\_Dataset obtained from the ILASH\_Heu runs on MTFL and CelebA. Similarly for the Taskonomy Dataset, when we are searching for the model for Taskonomy-1, we train the Auto-ILASH model on the two merged ILASH\_Dataset obtained from the ILASH\_Heu runs on Taskonomy-2 and Taskonomy-3. When training the Auto-ILASH model, we split the data in a 7:3 ratio for training and validation purposes respectively.

\begin{figure}[!ht]
\centering
\includegraphics[width=0.99\linewidth, height=0.4\textwidth]{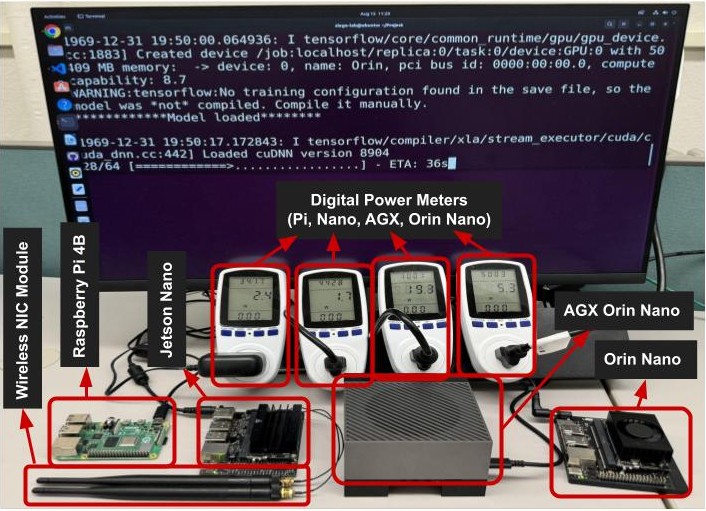}
\caption{Inferencing setup using target edge devices.}
\label{fig:inference}
\end{figure}
% \vspace{-0.2in}
\begin{table}[ht]
\caption{Efficacy of different machine learning algorithms for building the ILASH-NAS branching model. RF = Random Forest, DT = Decision Tree, Linear = Linear Regression, SVR = Support Vector Regression, and GrandBoost = Gradient Boosting Regressor.}
\centering
\renewcommand{\arraystretch}{1.3}
\small\addtolength{\tabcolsep}{4pt}
\begin{tabular}{ccccc}
\hline
\textbf{Model}     & \textbf{MAE}    & \textbf{MSE}     & \textbf{RMSE}   & \textbf{$R^2$}    \\ \hline
\textbf{RF}        & 0.033& 0.002& 0.045& 0.971  \\ 
\textbf{DT}        & 0.027  & 0.002& 0.044& 0.973  \\
\textbf{Linear}    & 0.054& 0.004& 0.063& 0.942  \\ 
\textbf{SVR}       & 0.113  & 0.020& 0.142& 0.701  \\ 
\textbf{GradBoost} & 0.040& 0.002& 0.049& 0.965\\ \hline
\end{tabular}
\label{tab:ml-model}
% \vspace{-0.2in}
\end{table}
\begin{figure*}[!t]
\centering
\includegraphics[width=0.8\linewidth]{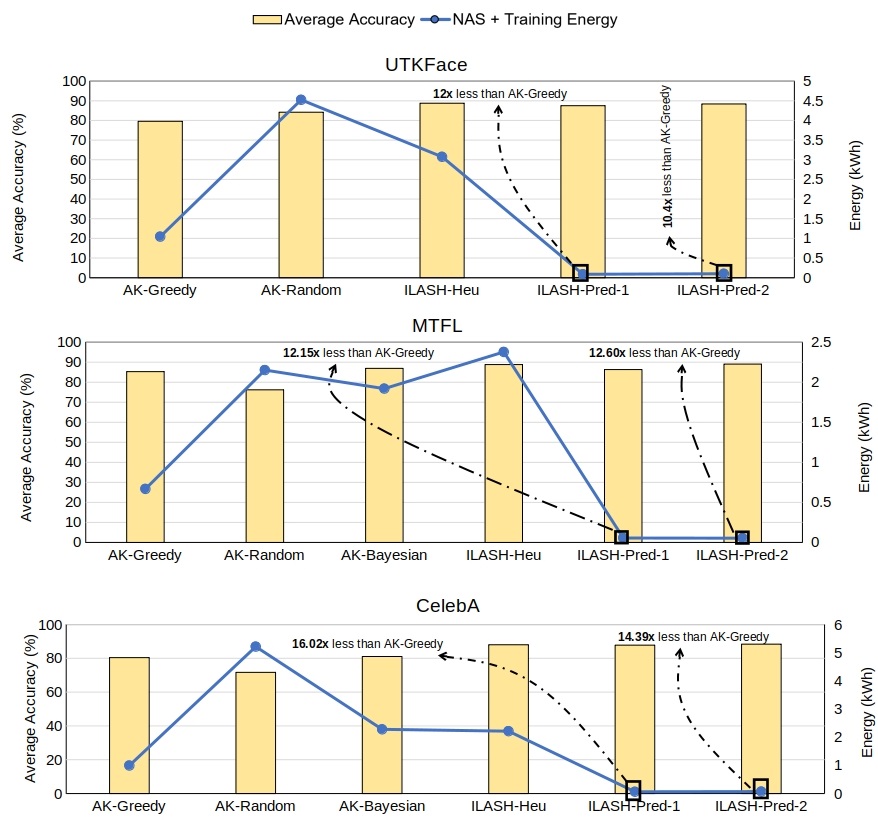}
\caption{\textbf{UTKFace, MTFL, and CelebA Datasets:} Neural search efficiency and model accuracy comparison between ILASH and other state-of-the-art multitask NAS algorithms.}
\label{fig:res_utk_celeb}
% \vspace{-0.2in}
\end{figure*}

% ILASH\_Heu datasets from 

% Also, the Auto-ILASH model when used for searching the network using UTKFace

\subsection{Experimental Setup: Edge Devices}

As target edge devices (for measuring inference-time efficiency) we have used NVIDIA Jetson AGX Orin, Orin Nano, Jetson Nano, and Raspberry Pi 4 Model B. For these devices, we measure the power consumption using plug-in monitors and calculate the equivalent energy consumption (KWh-PUE) and \ce{CO2} emissions with the help of eqn. ~\ref{eqn1} and \ref{eqn2}. Our edge computing setup is shown in Fig.~\ref{fig:inference}). 

\subsection{Experimental Setup: Task Models}
For both process (ILASH-Heu, ILASH-Pred), when training the ILASH task models, we use Adam optimizer with a learning rate of $1\times10^{-3}$, batch size of 32, and 50 epochs. In the preprocessing stage, we resize all images to (128 × 128) and split them up into training, validation and testing sets with a ratio of 70\%, 10\% and 20\% respectively. For searching with Auto-Keras, we use a $max\_trial$ of 20 for all methods (Greedy, Random and Bayesian). All neural search and training are performed on the following setup: \{AMD EPYC 2 Rome CPU, Nvidia A100 GPU 40 GB, 128 GB RAM\}.
To measure the power draw of GPU and CPU during network search/training, we used pyJoules~\cite{powerapi2024pyjoules} (Python Library). We used eqn. \ref{eqn1} and \ref{eqn2} to calculate the power consumption and equivalent $CO_2$ emission for all experiments. We use MobileNet \cite{howard2017mobilenets} as the base model ($Model\_{base}$ in Algo.~\ref{algo:Heuristic_Based_Search} and Algo.~\ref{algo:Predictive_Based_Search}) for classification/regression tasks (UTKFace, MTFL, and CelebA Datasets). We use MobileNet as the encoder, followed by six Conv2DTranspose layers with channel sizes of 512, 256, 128, 64, 32, and 1 (with ReLU activation) acting as the decoder as the base model ($Model\_{base}$ in Algo.~\ref{algo:Heuristic_Based_Search} and Algo.~\ref{algo:Predictive_Based_Search}) for 2D semantic tasks (Taskonomy Datasets).

% and for 2D classification task we utilized MobileNet as the encoder, followed by six Conv2DTranspose layers with channel sizes of 512, 256, 128, 64, 32, and 1 (with ReLU activation) acting as the decoder.

%\input{tables/ilash_2d}

\subsection{Data Description}
\label{sec:dataDesc}
In this study, we considered three different datasets (a) UTKFace \cite{das2018mitigating}, (b) MTFL \cite{zhang2014facial} and (c) CelebA \cite{liu2018large} for image classification and regression tasks, and the Taskonomy \cite{zamir2018taskonomy} dataset for 2D semantic tasks. 

\subsubsection{\textbf{UTKFace}}
The UTKFace dataset is a widely used and publicly available collection of facial images that is often used in the field of computer vision and machine learning research.
% , specifically for problems related to the classification of age, race, and gender. 
The dataset consists of more than 20,000 facial images representing different individuals. 
Each image is annotated with the individual’s gender ($C_1$), race ($C_2$), and age ($R_1$) that are used as labels for our multi-task applications. The order in which each of these tasks are presented to the ILASH-NAS algorithms can affect the overall performance hence we report the ILSH\_Pred results for two task orders: (1) $C_1, C_2, R_1$ represented by ILSH\_Pred-1 in the results; (2) $C_2, R_1, C_1$ represented by ILSH\_Pred-2 in the results. 
% ILSH\_Pred-1 represents  $C_1, C_2, R_1$ task order for  and $C_2, R_1, C_1$ task order for ILSH\_Pred-2

% Age is often quantified using numerical values, whereas gender is completely categorized as either male or female. Race, on the other hand, encompasses several classifications such as White, Black, Asian, Indian, and other diverse categories (not considered in this study).

\begin{table*}[!h]
\centering
% \scriptsize\addtolength{\tabcolsep}{-2pt}
\caption{\textbf{UTKFace, MTFL, and CelebA Datasets:} Results showing inferencing efficacy of ILASH and Autokeras (AK) on NVIDIA Jetson AGX Orin, Orin Nano, Jetson Nano, and Raspberry Pi.}
\label{tab:com_pi_Nano}
\renewcommand{\arraystretch}{1.3}
\scriptsize\addtolength{\tabcolsep}{-3pt}
\begin{tabular}{l|c|ccc|ccc|ccc|ccc}
\hline
\multirow{4}{*}{\textbf{Datasets}} & \multirow{4}{*}{\textbf{Methood}} & \multicolumn{3}{c|}{\textbf{\begin{tabular}[c]{@{}c@{}}Inference in Jetson AGX Nano \\ (per image)\end{tabular}}} 
& \multicolumn{3}{c|}{\textbf{\begin{tabular}[c]{@{}c@{}}Inference in Jetson Orin Nano \\ (per image) \end{tabular}}}
&
\multicolumn{3}{c|}{\textbf{\begin{tabular}[c]{@{}c@{}}Inference in  Jetson Nano \\ (per image) \end{tabular}}} 
&
\multicolumn{3}{c}{\textbf{\begin{tabular}[c]{@{}c@{}}Inference in Raspberry Pi \\ (per image) \end{tabular}}} 
\\ \cline{3-14}

&  & {\begin{tabular}[c]{@{}c@{}}KWh-PUE \\ ($\times10^{-8}$) \end{tabular}}  & {\begin{tabular}[c]{@{}c@{}}\ce{CO2} \\ Emission\\ ($\times10^{-8}$ lbs)\end{tabular}}  & FPS & \begin{tabular}[c]{@{}c@{}}KWh-PUE \\ ($\times10^{-8}$) \end{tabular} & {\begin{tabular}[c]{@{}c@{}}\ce{CO2} \\ Emission\\ ($\times10^{-8}$ lbs)\end{tabular}}  & FPS 
 & \begin{tabular}[c]{@{}c@{}}KWh-PUE \\ ($\times10^{-6}$) \end{tabular} & {\begin{tabular}[c]{@{}c@{}}\ce{CO2} \\ Emission\\ ($\times10^{-7}$ lbs)\end{tabular}}  & FPS 
  & \begin{tabular}[c]{@{}c@{}}KWh-PUE \\ ($\times10^{-7}$) \end{tabular} & {\begin{tabular}[c]{@{}c@{}}\ce{CO2} \\ Emission\\ ($\times10^{-7}$ lbs)\end{tabular}}  & FPS 
\\ \hline

\multirow{5}{*}{\begin{tabular}[c]{@{}c@{}}UTKFace \\ \cite{das2018mitigating}\end{tabular}}
 & AK-Greedy & $3.45$ & $3.29$ & 131.33 & $2.95$ & $2.81$ & 112.26 & 1.16 & 1.11 & 15.95 & 2.31 & 2.20 & 5.31 \\ %\cline{2-14} 
 &  AK-Random & 1.33 & 1.27 & 161.46 & 2.05 & 1.95 & 151.86 & 1.17 & 1.06 & 18.56 & 2.11 & 2.02 & 6.43
 \\
 %\cline{2-14} 
  &  ILASH-Heu & 1.0 & 0.957 & 351.69 & 1.58 & 1.46 & 288.56 & 0.597 & 0.569 & 58.07 & 0.804 & 0.767 & 17.53 
  \\
 %\cline{2-14} 
  &  ILASH-Pred-1 & 1.06 & 1.01 & 357.02 &1.25 & 1.19 & 296.57 & 0.564 & 0.538 & 56.18 & 0.827 & 0.789 & 16.99
  \\
 %\cline{2-14} 
  &  ILASH-Pred-2 & 1.05 & 1.01 & 349.85 & 1.01 & 0.965 & 291.62 & 0.481 & 0.458 & 55.48 & 0.774 & 0.738 & 18.45
 \\ \hline

\rowcolor[HTML]{EFEFEF} \cellcolor[HTML]{EFEFEF} & AK-Greedy & 5.37 & 5.13 & 92.17 & 4.63 & 4.42 & 83.56 & 5.18 & 4.94 & 10.56 & 5.24 & 5.0 & 2.59
  \\
 %\cline{2-14}
\rowcolor[HTML]{EFEFEF} \cellcolor[HTML]{EFEFEF}  &  AK-Bayesian & 6.49 & 6.19 & 66.91 & 6.02 & 5.74 & 57.62 & 5.24 & 5.0 & 11.48 & 4.99 & 4.76 & 2.55
  \\ %\cline{2-14}

\rowcolor[HTML]{EFEFEF} \cellcolor[HTML]{EFEFEF} &  AK-Random & 2.34 & 2.23 & 145.16 & 2.45 & 2.34 & 129.78 & 4.32 & 4.12 & 21.56 & 2.33 & 2.23 & 7.12
  \\
 %\cline{2-14} 

\rowcolor[HTML]{EFEFEF} \cellcolor[HTML]{EFEFEF} &  ILASH-Heu & 1.04 & 0.993 & 256.18 & 0.107 & 0.102 & 221.56 & 0.682 & 0.651 & 52.12 & 0.802 & 0.766 & 17.97
 \\
% \cline{2-14} 

\rowcolor[HTML]{EFEFEF} \cellcolor[HTML]{EFEFEF} &  ILASH-Pred-1 & 0.830 & 0.791 & 265.68 & 0.937 & 0.894 & 228.04 & 0.689 & 0.657 & 54.15 & 0.649 & 0.619 & 23.71
  
  \\
% \cline{2-14} 

\rowcolor[HTML]{EFEFEF} \multirow{-5}{*}{\cellcolor[HTML]{EFEFEF}\begin{tabular}[c]{@{}c@{}}MTFL \\ \cite{zhang2014facial}\end{tabular}} &  ILASH-Pred-2 & 0.913 & 0.871 & 267.66 & 0.935 & 0.892 & 184.7 & 0.715 & 0.682 & 53.87 & 0.576 & 0.550 & 24.15
 \\ \hline

 \multirow{5}{*}{\begin{tabular}[c]{@{}c@{}}CelebA \\ \cite{liu2018large}\end{tabular}}
 & AK-Greedy & 2.04 & 1.94 & 214.6 & 2.44 & 2.33 & 186.21 & 1.90 & 1.81 & 15.86 & 2.37 & 2.27 & 6.47
 
  \\
 %\cline{2-14}
 
  &  AK-Bayesian & 5.90 & 5.62 & 56.63 & 6.42 & 6.13 & 53.55 & 2.11 & 2.01 & 9.4 & 7.17 & 6.84 & 2.02 
  
  \\
 %\cline{2-14} 
 &  AK-Random & 5.49 & 5.23 & 56.7 & 6.48 & 6.18 & 52.22 & 1.85 & 1.76 & 11.5 & 7.15 & 6.83 & 2.03
 
 \\
 %\cline{2-14} 
  &  ILASH-Heu & 8.31 & 0.793 & 337.76 & 0.814 & 0.776 & 268.79 & 0.579 & 0.553 & 59.06 & 0.599 & 0.571 & 23.54
  
  \\
 %\cline{2-14} 
 &  ILASH-Pred-1 & 0.735 & 0.701 & 340.36 & 0.806 & 0.769 & 275.56 & 0.543 & 0.518 & 58.14 & 0.412 & 0.393 & 23.56 \\
  %\cline{2-14} 

&  ILASH-Pred-2 & 0.836 & 0.798 & 336.48 & 0.681 & 0.650 & 296.42 & 0.511 & 0.488 & 62.15 & 3.88 & 3.70 & 25.46
 \\ \hline 

\end{tabular}
\vspace{-0.2in}
\end{table*}

\subsubsection{\textbf{MTFL}}
The multi-task facial landmark (MTFL) dataset is a collection of facial images that have been annotated with multiple sets of facial landmarks/attributes. These landmarks are defined as crucial points or coordinates on the face that correspond to distinct facial features, including but not limited to the eyes, nose, and mouth. The dataset includes attributes (used as multi-task labels) such as gender ($C_1$), smiling expression ($C_2$), glasses-wearing ($C_3$), and head pose ($C_4$). The order in which each of these tasks are presented to the ILASH-NAS algorithms can affect the overall performance hence we report the ILSH\_Pred results for two task orders: (1) $C_1, C_2, C_3, C_4$ represented by ILSH\_Pred-1 in the results; (2) $C_3, C_4, C_1, C_2$ represented by ILSH\_Pred-2. 
% ILSH\_Pred-1 represents  $C_1, C_2, R_1$ task order for  and $C_2, R_1, C_1$ task order for ILSH\_Pred-2
% They are widely used in the field of the computer vision and facial analysis. 
% We have annotated the attributes as: gender($C_1$), smiling ($C_2$), wearing glasses ($C_3$), and head pose ($C_4$). 
% We used $C_1, C_2, C_3, C_4$ task order for ILASH\_Pred-1 and $C_3, C_4, C_1, C_2$ task order for ILASH\_Pred-2.

\subsubsection{\textbf{CelebA}}
CelebA is widely used for face recognition, detection, and facial landmark prediction. We consider 5 attributes i.e., Gender ($C_1$), Hair color ($C_2$), Hair style ($C_3$), Smiling ($C_4$) and Nose ($C_5$) for our multi-task classification application. We randomly choose 20K images for our experiments. The order in which each of these tasks are presented to the ILASH-NAS algorithms can affect the overall performance hence we report the ILSH\_Pred results for two task orders: (1) $C_1, C_2, C_3, C_4, C_5$ represented by ILSH\_Pred-1 in the results; (2) $C_4, C_2, C_5, C_3, C_1$ represented by ILSH\_Pred-2. 

% We used $C_1, C_2, C_3, C_4, C_5$ task order for ILASH\_Pred-1 and $C_4, C_2, C_5, C_3, C_1$ task order for ILASH\_Pred-2.

\subsubsection{\textbf{Taskonomy}}
Taskonomy is a widely used dataset for 2D semantic tasks. This dataset consists of 4.6 million indoor scenes from 537 buildings. It has 4 different versions: Tiny, Medium, Full and Full+ containing indoor images from 35, 138, 482 and 537 buildings respectively.
In this study, we use data from the tiny category to perform 2D edge texture ($T_1$), surface normal ($T_2$), and reshading ($T_3$) tasks. We divide the tiny taskonomy dataset into three parts based on unique buildings and every part is considered as a separate dataset to evaluate our approach. The order in which each of these tasks are presented to the ILASH-NAS algorithms can affect the overall performance hence we report the ILSH\_Pred results for two task orders: (1) $T_1, T_2, T_3$ represented by ILSH\_Pred-1 in the results; (2) $T_2, T_3, T_1$ represented by ILSH\_Pred-2 in the results. 

% We used 2D edge texture, surface normal, and reshading task order for ILASH\_Pred-1 and surface normal, reshading, 2D edge texture tasks order for ILASH\_Pred-2.

\section{Experimental Results and Analysis}
In this section, we will discuss the experimental results.

\begin{table*}[!h]
\centering
% \scriptsize\addtolength{\tabcolsep}{-2pt}
\caption{\textbf{Taskonomy Dataset:} Results showing inferencing efficacy of ILASH on NVIDIA Jetson AGX Orin, Orin Nano, Jetson Nano, and Raspberry Pi.}
\label{tab:com_pi_nano2}
\renewcommand{\arraystretch}{1.3}
\scriptsize\addtolength{\tabcolsep}{-3pt}
\begin{tabular}{c|c|ccc|ccc|ccc|ccc}
\hline
\multirow{4}{*}{\textbf{Datasets}} & \multirow{4}{*}{\textbf{Methood}} &  \multicolumn{3}{c|}{\textbf{\begin{tabular}[c]{@{}c@{}}Inference in Jetson AGX Nano \\ (per image) \end{tabular}}} 
& \multicolumn{3}{c|}{\textbf{\begin{tabular}[c]{@{}c@{}}Inference in Jetson Orin Nano \\ (per image)\end{tabular}}}
&
\multicolumn{3}{c|}{\textbf{\begin{tabular}[c]{@{}c@{}}Inference in  Jetson Nano \\ (per image) \end{tabular}}} 
&
\multicolumn{3}{c}{\textbf{\begin{tabular}[c]{@{}c@{}}Inference in Raspberry Pi \\ (per image) \end{tabular}}} 
\\ \cline{3-14}

& & {\begin{tabular}[c]{@{}c@{}}KWh-PUE \\ ($\times10^{-8}$) \end{tabular}}  & {\begin{tabular}[c]{@{}c@{}}\ce{CO2} \\ Emission\\ ($\times10^{-8}$ lbs)\end{tabular}}  & FPS & \begin{tabular}[c]{@{}c@{}}KWh-PUE \\ ($\times10^{-8}$) \end{tabular} & {\begin{tabular}[c]{@{}c@{}}\ce{CO2} \\ Emission\\ ($\times10^{-8}$ lbs)\end{tabular}}  & FPS 
 & \begin{tabular}[c]{@{}c@{}}KWh-PUE \\ ($\times10^{-6}$) \end{tabular} & {\begin{tabular}[c]{@{}c@{}}\ce{CO2} \\ Emission\\ ($\times10^{-7}$ lbs)\end{tabular}}  & FPS 
  & \begin{tabular}[c]{@{}c@{}}KWh-PUE \\ ($\times10^{-7}$) \end{tabular} & {\begin{tabular}[c]{@{}c@{}}\ce{CO2} \\ Emission\\ ($\times10^{-7}$ lbs)\end{tabular}}  & FPS 
\\ \hline

\multirow{3}{*} 1
  &  ILASH-Heu & 1.0 & 0.957 & 351.69 & 1.58 & 1.46 & 288.56 & 0.597 & 0.569 & 58.07 & 0.804 & 0.767 & 17.53 
  \\
 %\cline{2-14} 
  &  ILASH-Pred-1 & 1.06 & 1.01 & 357.02 &1.25 & 1.19 & 296.57 & 0.564 & 0.538 & 56.18 & 0.827 & 0.789 & 16.99
  \\
% \cline{2-14} 
  &  ILASH-Pred-2 & 1.05 & 1.01 & 349.85 & 1.01 & 0.965 & 291.62 & 0.481 & 0.458 & 55.48 & 0.774 & 0.738 & 18.45
 \\ \hline

\rowcolor[HTML]{EFEFEF} \cellcolor[HTML]{EFEFEF} &  ILASH-Heu &  1.04 & 0.993 & 256.18 & 0.107 & 0.102 & 221.56 & 0.682 & 0.651 & 52.12 & 0.802 & 0.766 & 17.97
 \\
 %\cline{2-14} 
\rowcolor[HTML]{EFEFEF} \cellcolor[HTML]{EFEFEF}  &  ILASH-Pred-1  &  0.830 & 0.791 & 265.68 & 0.937 & 0.894 & 228.04 & 0.689 & 0.657 & 54.15 & 0.649 & 0.619 & 23.71
  
  \\
 %\cline{2-14} 
\rowcolor[HTML]{EFEFEF} \multirow{-3}{*}{\cellcolor[HTML]{EFEFEF}2} &  ILASH-Pred-2 & 0.913 & 0.871 & 267.66 & 0.935 & 0.892 & 184.7 & 0.715 & 0.682 & 53.87 & 0.576 & 0.550 & 24.15
 \\ \hline

 \multirow{3}{*} 3
 
  &  ILASH-Heu &  8.31 & 0.793 & 337.76 & 0.814 & 0.776 & 268.79 & 0.579 & 0.553 & 59.06 & 0.599 & 0.571 & 23.54
  
  \\
 %\cline{2-14} 
 &  ILASH-Pred-1 & 0.735 & 0.701 & 340.36 & 0.806 & 0.769 & 275.56 & 0.543 & 0.518 & 58.14 & 0.412 & 0.393 & 23.56 \\
 % \cline{2-14} 

&  ILASH-Pred-2 & 0.836 & 0.798 & 336.48 & 0.681 & 0.650 & 296.42 & 0.511 & 0.488 & 62.15 & 3.88 & 3.70 & 25.46
 \\ \hline 

\end{tabular}
% \vspace{-0.2in}
\end{table*}
\begin{table}[ht]
\renewcommand{\arraystretch}{1.3}
\scriptsize\addtolength{\tabcolsep}{-1pt}
\centering
\caption{\textbf{Taskonomy Dataset:} Neural search efficiency and model MAE (performance) comparison between different ILASH settings.}
\label{tab:2d_runtime}
\begin{tabular}{c|c|c|ccc}
\hline
\multirow{3}{*}{\textbf{Datasets}} &
\multirow{3}{*}{\textbf{Method}} &
\multirow{3}{*}{\textbf{\begin{tabular}[c]{@{}c@{}}Average \\ 
    MAE \end{tabular}}}
 &
\multicolumn{3}{c} {\textbf{NAS and Final Training}}
\\ \cline{4-6}
 & & & 
\textbf{\begin{tabular}[c]{@{}c@{}}Runtime \\ 
    (Hrs) \end{tabular}} &
  \textbf{kWh-PUE} &
  \textbf{\begin{tabular}[c]{@{}c@{}}\ce{CO2} emission \\ (lbs)\end{tabular}} \\ \hline

\multirow{3}{*} 1  & ILASH-Heu & 0.0821 & 60.2 & 6.34 & 6.05 \\ %\cline{2-6}
 & ILASH-Pred-1  &  0.0749 & 1.58 & 0.287 & 0.273 \\ %\cline{2-6}
& ILASH-Pred-2  &  0.0756 &1.44 &0.247 & 0.245 \\ \hline

\rowcolor[HTML]{EFEFEF} \cellcolor[HTML]{EFEFEF}  & ILASH-Heu & 0.0798 & 48.73 & 5.73 & 5.44 \\ %\cline{2-6}
\rowcolor[HTML]{EFEFEF} \cellcolor[HTML]{EFEFEF}  & ILASH-Pred-1  & 0.0787 & 1.30 & 0.239 & 0.228 \\ %\cline{2-6}
\rowcolor[HTML]{EFEFEF}\multirow{-3}{*} {\cellcolor[HTML]{EFEFEF}2} & ILASH-Pred-2  & 0.0775 & 1.18 & 0.183 & 0.174 \\ \hline

\multirow{3}{*} 3  &  ILASH-Heu  & 0.0810 & 53.6 & 5.78 & 5.49 \\ %\cline{2-6}
 & ILASH-Pred-1  & 0.0785 & 1.48 & 0.262 & 0.249 \\ %\cline{2-6}
& ILASH-Pred-2  &  0.0795 & 1.39 & 0.240 & 0.228 \\ \hline

\end{tabular}
\vspace{-0.2in}
\end{table}

\subsection{Auto-ILASH Model Selection}
The Auto-ILASH model used in Algorithm~\ref{algo:Predictive_Based_Search} (ILASH-Pred, the predictive version of ILASH-NAS) is trained on the data, ILASH\_Dataset, obtained from Algorithm~\ref{algo:Heuristic_Based_Search} runs (ILASH-Heu, the heuristics version of ILASH-NAS). We have experimented with different machine learning (ML) algorithms for creating the most optimal Auto-ILASH model. Table \ref{tab:ml-model} shows these experimental results. Among the five ML model, Decision Tree performs better in terms of MAE, MSE, RMSE and $R^2$ score. Based on this evaluation we decided to use the Decision Tree algorithm for building the final Auto-ILASH model.
% to predict the goodness of branched layer shared model.

%\input{tables/ILASH}

\subsection{Evaluation on the UTKFace, MTFL, and CelebA Datasets}
In Figure~\ref{fig:res_utk_celeb}, we compare the final model accuracy (across all the tasks for the given dataset) and the energy utilization for neural architecture search (+training). We see that the overall final accuracy of the system remains mostly stable across all the techniques, however, our proposed ILASH-Pred technique (for both task ordering) is extremely energy efficient. The order in which each of these tasks are presented to the ILASH-NAS algorithms can affect the overall performance hence we report the ILSH\_Pred results for two task orders (more details in Section~\ref{sec:res}). Here AK-Greedy refers to AutoKeras with `greedy' tuner, AK-Random refers to AutoKeras with `random' tuner, and AK-Bayesian refers to AutoKeras with `bayesian' tuner. For more details about AutoKeras please refer to \cite{jin2019auto, JMLR:v24:20-1355}. 
We see very similar trend across all the three datasets MTFL, CelebA, and UTKFace. AK-Greedy is the least energy intensive approach among all AutoKeras techniques. Even then ILSH\_Pred is 10-16 times more efficient. This is made possible due to the Auto\_ILASH model making the branching decisions with very little delay. ILASH\_Heu is not the most optimal approach and should be only used for creating the initial dataset necessary to build the Auto\_ILASH model for ILASH\_Pred.

Next we evaluate the efficacy (during inferencing/deployment) of the final searched models obtained via different variants of AutoKeras and ILASH-NAS for the UTKFace, MTFL, and CelebA Datasets (see Table~\ref{tab:com_pi_Nano}). We use four different edge devices (NVIDIA Jetson AGX Orin, Orin Nano, Jetson Nano, and Raspberry Pi 4 Model B) to extensively test the efficacy of the final model. These results conclusively demonstrates that both ILASH\_Heu and ILASH\_Pred leads to more optimal neural networks for edge devices in terms of energy consumption (KWh-PUE), \ce{CO2} emission, and frames processed per second (FPS).

\subsection{Evaluation on the Taskonomy Dataset}
Next we perform a separate set of experiments to estimate the efficacy of ILASH-NAS (both ILASH\_Heu and ILASH\_Pred) for building multi-task networks to support 2D semantics tasks (2D edge texture, surface normal, and reshading). More details about the dataset used can be found in Section~\ref{sec:dataDesc}. Table~\ref{tab:2d_runtime} shows the corresponding experimental results. Note that, we are unable to compare our results with AutoKeras because it (AutoKeras) does not support 2D semantics tasks. Hence, to the best of our knowledge, ILASH-NAS is the only multi-task neural architecture search framework that supports 2D semantics tasks. In Table~\ref{tab:2d_runtime} we observe that ILASH\_Pred (utilizing the Auto-ILASH ML Model) is approximately 38 times faster than ILASH-Heu method during the network search process resulting in nearly 22 times more energy efficiency (without much loss in final model MAE/performance). Next we evaluate the efficacy (during inferencing/deployment) of the final searched models obtained via different variants of ILASH-NAS for the Taskonomy Dataset (see Table~\ref{tab:com_pi_nano2}). ILASH-Pred shows promising performance in every aspects.

% We observed same trend for datasets 1 and 2. 
% Table~\ref{tab:2d_runtime} shows the experimental results for 2D multi-classification task. We can see that for dataset 1, ILASH-Heu took 60.2 hours resulting 6.34 kWh-PUE energy consumption for NAS and final training with average MAE of 0.0821. Where ILASH-Pred-1 and ILASH-Pred-2 took 1.58 and 1.44 hours respectively with 0.0749 and 0.0756 average MAE score. That means predictive ILASH is approximately 38x faster than ILASH-Heu method to search the optimal network resulting nearly 22x energy efficient while providing good performance. We observed same trend for datasets 1 and 2. 
% Table~\ref{tab:com_pi_nano2} shows the inferencing results for multi-classification task for 2D. ILASH-Pred-1 and ILASH-Pred-2 show promising performance in every aspects.
\section{Conclusion}
AI will play a major role in every aspect of human life in the near future. However, to make the pervasive AI dream a reality, we must analyze and mitigate its impact on energy utilization and the environment. In this work, we have presented a layer-shared neural architecture and a predictive NAS algorithm that leads to a dramatic reduction in energy/emission overheads (during architecture search \& inferencing) for multi-task applications. Future works will explore AI-guided NAS for more complex applications and neural network topology.

\bibliographystyle{ieeetr}
\bibliography{bibfile}

\end{document}